\def\paperID{8149}
\def\confName{CVPR}
\def\confYear{2025}

\def\authorBlock{
  An Li\textsuperscript{*} \quad 
  Zhe Zhu\textsuperscript{*} \quad 
  Mingqiang Wei\textsuperscript{†} \\
  Nanjing University of Aeronautics and Astronautics \\
  {\tt\small \{lian, zhuzhe0619, mqwei\}@nuaa.edu.cn}
}

\newif\ifreview 
\newif\ifarxiv \newcommand{\arxiv}{\arxivtrue}
\newif\ifcamera 
\newif\ifrebuttal 

\arxiv % \review OR \arxiv OR \cameraready

\pdfoutput=1
\documentclass[10pt,twocolumn,letterpaper]{article}
\ifreview \usepackage[review]{cvpr} \fi
\ifarxiv \usepackage[pagenumbers]{cvpr} \fi
\ifrebuttal \usepackage[rebuttal]{cvpr} \fi
\ifcamera \usepackage{cvpr} \fi

\usepackage{graphicx}	
\usepackage{amsmath}	
\usepackage{amssymb}	
\usepackage{booktabs}
\usepackage{times}
\usepackage{microtype}
\usepackage{epsfig}
\usepackage[table,xcdraw,dvipsnames]{xcolor}
\usepackage{caption}
\usepackage{float}
\usepackage{placeins}
\usepackage{color, colortbl}
\usepackage{stfloats}
\usepackage{enumitem}
\usepackage{tabularx}
\usepackage{xstring}
\usepackage{multirow}
\usepackage{xspace}
\usepackage{url}
\usepackage{subcaption}
\usepackage{xcolor}
\usepackage[hang,flushmargin]{footmisc}
\usepackage{CJKutf8}
\ifcamera \usepackage[accsupp]{axessibility} \fi

% ...

\ifarxiv  \fi

\newcommand{\R}[1]{{%
    \textbf{%
        \ifstrequal{#1}{1}{\textcolor{red}{R#1}}{%
        \ifstrequal{#1}{2}{\textcolor{blue}{R#1}}{%
        \ifstrequal{#1}{3}{\textcolor{magenta}{R#1}}{%
        \ifstrequal{#1}{4}{\textcolor{teal}{R#1}}{%
                           \textcolor{cyan}{R#1}%
        }}}}%
    }%
}}
\usepackage{xr-hyper}

\makeatletter
\newcommand*{\addFileDependency}[1]{
  \typeout{(#1)}
  \@addtofilelist{#1}
  \IfFileExists{#1}{}{\typeout{No file #1.}}
}

\makeatother
\newcommand*{\myexternaldocument}[1]{
    \externaldocument{#1}
    \addFileDependency{#1.tex}
    \addFileDependency{#1.aux}
}

\definecolor{cvprblue}{rgb}{0.21,0.49,0.74}
\usepackage[pagebackref,breaklinks,colorlinks,citecolor=cvprblue]{hyperref}
\usepackage[capitalize]{cleveref}
\crefname{section}{Sec.}{Secs.}
\crefname{table}{Table}{Tables}
\crefname{figure}{Fig.}{Figs.}

\ifarxiv \crefname{appendix}{App.}{Apps.}
\else \crefname{appendix}{Suppl.}{Suppls.} \fi

\unless\ifarxiv \myexternaldocument{_supplementary} \fi

\begin{document}
%% TITLE
\title{GenPC: Zero-shot Point Cloud Completion via 3D Generative Priors}
\author{\authorBlock}
\maketitle
\renewcommand{\thefootnote}{}
\footnotetext{\textsuperscript{*} Equal Contribution \textsuperscript{†} Corresponding author}
\begin{abstract}
% Abstract goes here.
Existing point cloud completion methods, which typically depend on predefined synthetic training datasets, encounter significant challenges when applied to out-of-distribution, real-world scans.
To overcome this limitation, we introduce a zero-shot completion framework, termed GenPC, designed to reconstruct high-quality real-world scans by leveraging explicit 3D generative priors.
Our key insight is that recent feed-forward 3D generative models, trained on extensive internet-scale data, have demonstrated the ability to perform 3D generation from single-view images in a zero-shot setting.
To harness this for completion, we first develop a Depth Prompting module that links partial point clouds with image-to-3D generative models by leveraging depth images as a stepping stone. To retain the original partial structure in the final results, we design the Geometric Preserving Fusion module that aligns the generated shape with input by adaptively adjusting its pose and scale.
Extensive experiments on widely used benchmarks validate the superiority and generalizability of our approach, bringing us a step closer to robust real-world scan completion.
\end{abstract}
\section{Introduction}
\label{sec:intro}
Point clouds, as an essential form of 3D representation, are widely used in various applications. However, due to factors such as self-occlusion, camera viewpoint limitations, and sensor resolution, the acquired point clouds are often incomplete. This issue significantly hinders downstream tasks. Therefore, developing effective and robust methods for completing real-world partial point clouds is crucial for achieving a comprehensive understanding of the real world.

\begin{figure}
    \centering
    \includegraphics[width=1\linewidth]{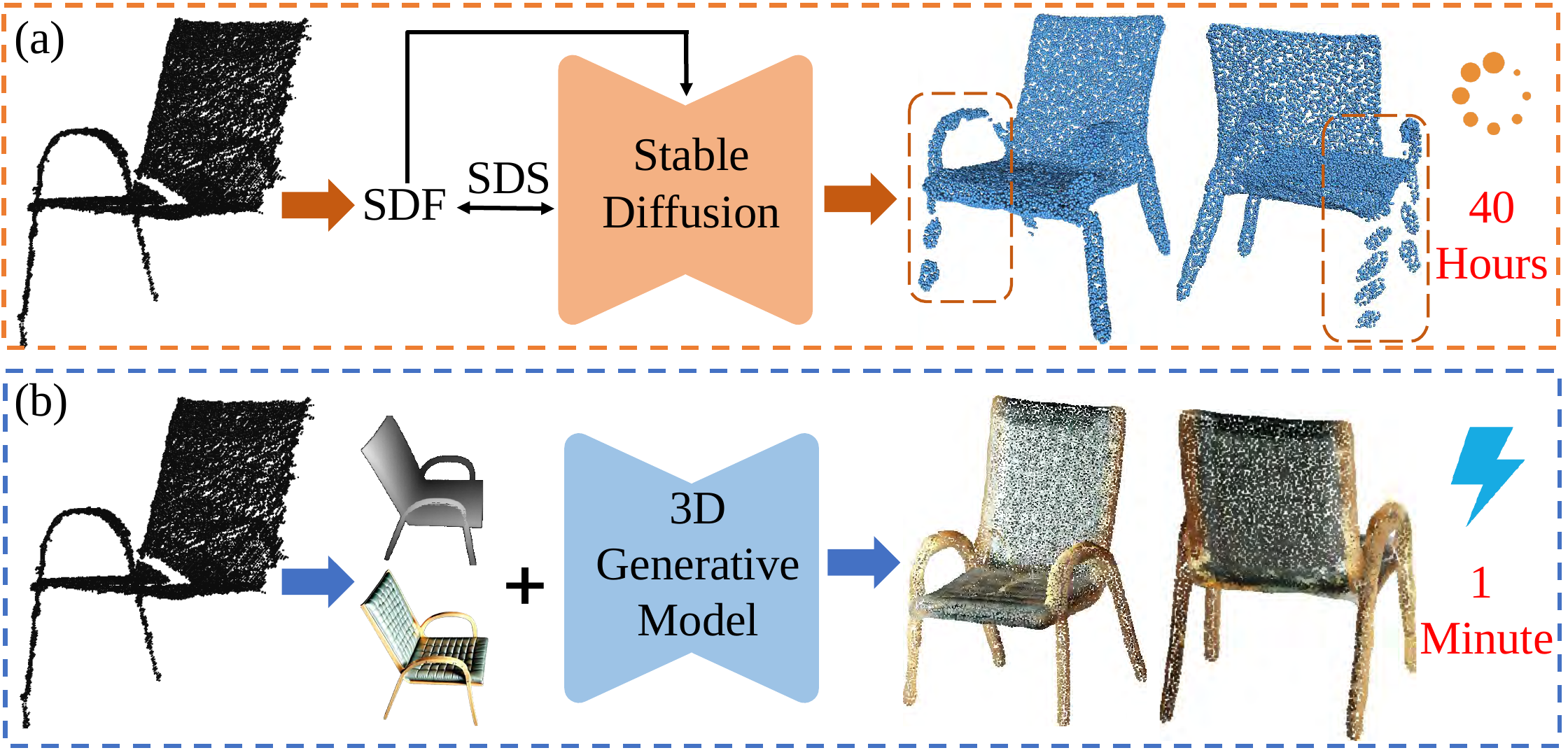}
    \caption{Difference between our GenPC with previous zero-shot point cloud completion method~\cite{sds-complete}. (a) SDS-Complete~\cite{sds-complete} uses the SDS loss to directly extract prior knowledge from a 2D diffusion model, featuring time-consuming optimization and suboptimal completion results. (b) The proposed GenPC leverages explicit priors provided by a 3D generative model, achieving improved completion quality with significantly reduced inference time.}
    \label{fig:teaser}
\end{figure}

In recent years, numerous deep learning-based point cloud completion methods~\cite{Pcn,Pointr,AdaPoinTr,pf-net,Snowflakenet,SVDFormer,Geoformer} have shown remarkable success. These approaches utilize carefully designed neural networks to extract shape patterns from input point clouds, enabling them to generate detailed geometric structures to complete missing portions of the point cloud. 
% Some methods~\cite{zhang2021view,zhu2023csdn} additionally incorporate corresponding 2D information to aid in point cloud completion. 
Although these techniques perform well on trained or similar categories, they rely on labeled 3D training data and exhibit limited generalization to categories unseen during training. Moreover, constrained by domain gaps between synthetic training data and real-world scans, these models tend to perform poorly when applied to downstream tasks.

% 简单介绍两个zeroshot方法，没有用3D大模型导致效果有限，且test-time训练时间长
With the impressive zero-shot generation capabilities of pre-trained 2D diffusion models~\cite{stablediffusion}, numerous studies~\cite{DreamFusion,zero123,DreamGaussian} have emerged that utilize these models for 3D generation tasks. 
Enlightened by these successes, sds-complete~\cite{sds-complete} first utilized 2D priors for zero-shot shape completion. This method fits the input partial point cloud surface using Signed Distance Functions (SDF) and leverages Score Distillation Sampling (SDS)~\cite{DreamFusion} to extract 2D diffusion priors for completion. 
Later, Huang et al.~\cite{zeroshotpointcloudcompletion} proposed a similar SDS-based framework, but used 3D Gaussian splatting~\cite{3dgs} to initialize the partial point cloud as 3D Gaussians.
Although these methods demonstrate improved zero-shot completion capabilities compared to training-based counterparts, they are time-consuming, as they require training a radiance field from scratch for each incomplete point cloud. Additionally, SDS loss often leads to coarse geometric details, limiting their reconstructing quality.

Recently, scalable network architectures and large-scale 3D datasets have propelled the success of feed-forward 3D generative models~\cite{LRM,LGM,instantmesh}. Once trained, these models achieve impressive zero-shot generation quality within seconds. This raises an intriguing question: \textit{``Can we leverage this 3D generative capability for point cloud completion?''}
To answer it, we introduce a novel zero-shot point cloud completion framework, GenPC. Unlike previous 2D diffusion-based approaches~\cite{sds-complete,zeroshotpointcloudcompletion}, GenPC utilizes explicit 3D priors from an image-to-3D generative model to enhance zero-shot completion quality, significantly improving inference speed.

\begin{figure*}
    \centering
    \includegraphics[width=0.9\textwidth]{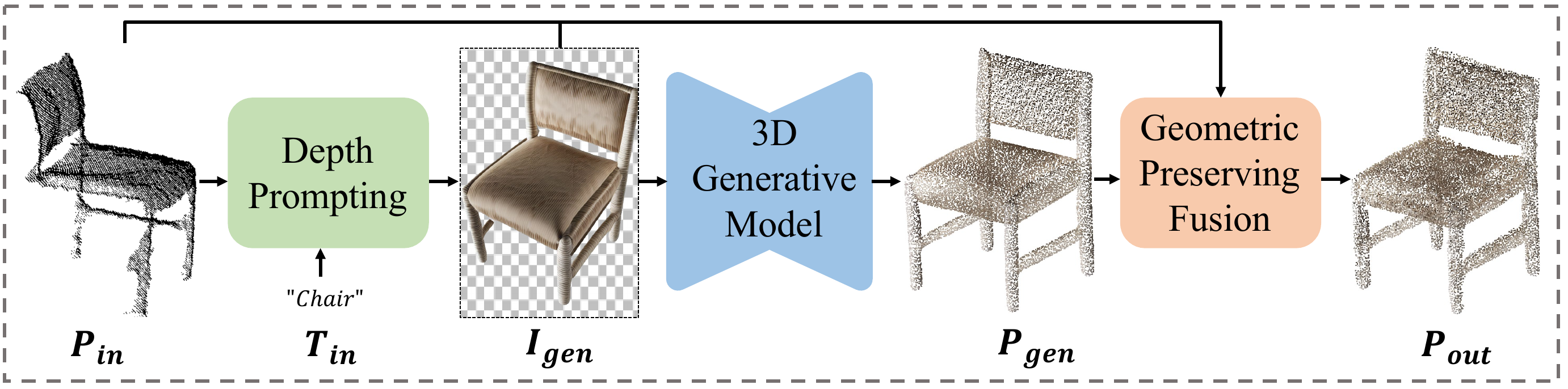}
    \caption{The architecture of GenPC. The Depth Prompting module first prompts the depth-guided 2D generative model with the partial input and generates an RGB image, which is fed into an image-to-3D generative model, producing a 3D shape. 
    The Geometric Preserving Fusion module then integrates the generated shape with the partial point cloud.} 
    \label{fig:pipeline}
\end{figure*}

%%%%%%%%%%%%%%%%%%%%%%%%%%%%%%%%%%%%%%%%%%%%%%%%%%%%%%
% 两个点（技术贡献）
% Therefore, we need a bridge to connect partial point clouds with the image-to-3D generation model, and this bridge is depth. By utilizing depth extracted from the partial input point cloud, we can effectively guide the generation process to ensure a high match between the generated 3D object and the partial point cloud, better leveraging 3D priors for precise control, thereby enhancing completion quality and reducing completion time.
As illustrated in Figure ~\ref{fig:pipeline}. To leverage the powerful zero-shot generation capabilities of the image-to-3D generative model, we first need an image input. To address the issue of partial point clouds not directly providing image input for these models, we introduce a Depth Prompting module. This module estimates the scanning viewpoint of the partial point cloud and extracts depth, effectively bridging the modality gap between the point cloud and the generative model.
After generating the 3D shape, a significant issue arises: the generated 3D shape may differ from the input partial point cloud in terms of scale, pose, and shape. To align it with the input point cloud and retain the original geometric structure, we introduce the Geometric Preserving Fusion module. This module first dynamically adjusts the scale and pose using a scaling factor at both geometric and semantic levels. 
In addition, we can further refine the point cloud using the SDS loss, minimizing shape detail discrepancies caused by multi-stage error accumulation.
By leveraging explicit geometric priors offered by the 3D generative model, our approach avoids the need for optimization from scratch, enabling faster inference and superior completion quality.

In summary, our contributions are as follows:
\begin{itemize}
    \item We design a novel zero-shot completion framework called GenPC, which significantly improves real-world scan completion by prompting a pre-trained 3D generative model.
    \item We propose a Depth Prompting module to bridge the modality gap between partial scans and generative models by utilizing depth images as a stepping stone.
    \item We introduce the novel Geometric Preserving Fusion module for refining the initial generated results. It adaptively aligns the generated content with partial input, ensuring that the final result is both semantically accurate and structurally faithful.
    \item Extensive experiments demonstrate that GenPC achieves state-of-the-art performance on real-world datasets while significantly reducing completion time.
\end{itemize}

\section{Related Work}
\label{sec:related}
\subsection{Point cloud completion}
% Recent years have seen significant progress in point cloud completion. 
Early methods ~\cite{3depn,han2017high,varley2017shape,xie2020grnet} primarily used voxels as intermediate representations and performed completion using 3D convolutions. 
% However, they are often limited by the resolution of the voxels. With the development of point-based networks like PointNet~\cite{qi2017pointnet},  PCN~\cite{Pcn} directly generates high-resolution complete point clouds in a coarse-to-fine manner.
However, they are often limited by the resolution of the voxels. With the development of point-based networks like PointNet~\cite{qi2017pointnet}, various point cloud tasks can be handled by end-to-end networks \cite{repcdnet,geometry,punet,pointclean,votenet}. Among them, PCN~\cite{Pcn} is the first work that directly generates high-resolution complete point clouds in a coarse-to-fine manner for point cloud completion.
A similar generation strategy is also adopted in a series of following works~\cite{liu2020morphing,pf-net,wang2020cascaded,sa-net,wen2021pmp}. Transformer~\cite{vaswani2017attention} has also been leveraged in recent works. PoinTr~\cite{Pointr} treats the point cloud as a token sequence, using transformer encoder-decoder to predict the missing parts. SnowflakeNet~\cite{Snowflakenet} designs a transformer decoder with skip connections to refine the point cloud. 
Another line of works~\cite{zhang2021view,zhu2023csdn} enhances completion performance using 2D information. 
% ViPC~\cite{zhang2021view} takes an additional 2D image as input to provide additional color information. Following it, CSDN~\cite{zhu2023csdn} integrates shape features from images into global features, generating finer geometric structures. 
Different from the above approaches, SVDFormer~\cite{SVDFormer} and GeoFormer~\cite{Geoformer} project point clouds into 2D depth images, requiring information from only partial input.

Although these methods perform well on synthetic datasets, their reliance on training data causes performance degradation on out-of-distribution real-world scans and previously unseen categories. Recent unsupervised~\cite{pcl2pcl,unsupervisedgan,xie2021stylegan,c4c} and self-supervised approaches~\cite{ppnetbmvc,aclspc,p2c} have alleviated this issue to some extent; however, the completion results remain suboptimal.
% Furthermore, due to the domain gap, these methods often experience a significant drop in performance when applied to real-world scanned data. Additionally, these methods tend to work well only for the specific categories or similar categories seen during training, lacking generalization to unseen categories. 
To address these limitations, SDS-Complete~\cite{sds-complete} formulates point cloud completion as a test-time optimization problem, introducing a zero-shot method that fits a Signed Distance Function (SDF) to the input partial point cloud. It leverages Score Distillation Sampling (SDS) to extract 2D priors from the Stable Diffusion~\cite{stablediffusion} model to complete the missing regions. Subsequently, Huang et al.\cite{zeroshotpointcloudcompletion} propose initializing the partial point cloud as 3D Gaussians and distilling prior knowledge from zero123\cite{zero123}. Although these methods exhibit impressive zero-shot completion capabilities, they require optimization from scratch for each incomplete point cloud, making them time-intensive. Moreover, reliance on implicit 2D diffusion priors limits the reconstruction of fine geometric details.
In this work, we leverage explicit priors from a pre-trained 3D generative model to enhance zero-shot point cloud completion quality while significantly reducing processing time.

\subsection{3D Generation}
DreamFusion~\cite{DreamFusion} is the first method to use 2D priors for 3D generation, introducing Score Distillation Sampling (SDS) to extract 2D priors from a pretrained diffusion model and guide the 3D generation process, inspiring numerous impressive works. Magic3D~\cite{magic3d} adopts DMTet~\cite{dmtet} as the 3D representation instead of NeRF~\cite{nerf} and then performs optimization using SDS. Fantasia3D~\cite{Fantasia3D} decouples the optimization of geometry and material properties. 
With the emergence of 3D Gaussian Splatting~\cite{3dgs}, a highly expressive 3D representation, the optimization time for 3D generation with SDS has been significantly reduced. DreamGaussian~\cite{DreamGaussian} firstly attempts to use SDS optimization for 3D Gaussians, reducing the optimization time to just a few minutes while achieving excellent results. GaussianDreamer~\cite{gaussiandreamer} initializes 3D Gaussians using point cloud priors, yielding impressive results. 
% 下面两段有点重复
% Despite the remarkable performance and generalization ability of these optimization-based methods, they still require several minutes of processing time.
Although the above methods are effective, they require several minutes or even hours for optimization. 
The emergence of large-scale datasets~\cite{objaverse,objaverseXL} has driven the development of faster feed-forward methods. Once trained, these methods can generate 3D objects within seconds through a single forward inference. Recently, LRM~\cite{LRM} demonstrated that a regression model can predict a NeRF from a single image within seconds. Based on this, InstantMesh~\cite{instantmesh} generates additional multi-view images from a single image and then reconstructs the mesh. However, both methods are limited by resolution. To address this, LGM~\cite{LGM} introduces an efficient representation of multi-view Gaussian features, enabling the prediction of high-resolution 3D Gaussian models. 

These feed-forward methods can generate high-quality 3D objects from a single image in a very short time while demonstrating strong generalization ability. 
We are motivated to leverage this advantage for point cloud completion, aiming to achieve superior zero-shot completion results while reducing optimization time.

\section{Method}
\label{sec:method}
The input of GenPC consists of a partial point cloud \( P_{in} \subseteq \mathbb{R}^{N \times 3} \) and a corresponding text prompt \( T_{in} \), where \( N \) represents the number of points in \( P_{in} \). Our goal is to obtain a complete shape $P_{out}$ that preserves the original structure in input.
As illustrated in Figure~\ref{fig:pipeline}, our method seamlessly incorporates an image-to-3D generative model into the point cloud completion process through the introduction of two innovative modules.
First, current image-to-3D models are designed to accept only 2D images as input. To adapt them for point cloud completion, we introduce the \textbf{Depth Prompting} module, which leverages depth images as a stepping stone to bridge the modality gap between partial point clouds and generative models.
After generating a 3D shape from the image-to-3D model, a key challenge arises: the original points in $P_{in}$ are not retained in the generated shape. To address this, we propose the \textbf{Geometric Preserving Fusion} module, which further aligns the initial generated shape with $P_{in}$, ensuring that the final result is both semantically accurate and structurally faithful.
% \textbf{2) Generation and Fusion}, which generates corresponding 3D objects, integrates it with the partial point cloud, and fills in the missing regions through the fusion process. \textbf{3) Optimization}, which reduces error accumulation and enhances completion quality.
\begin{figure}
    \centering
    \includegraphics[width=1\linewidth]{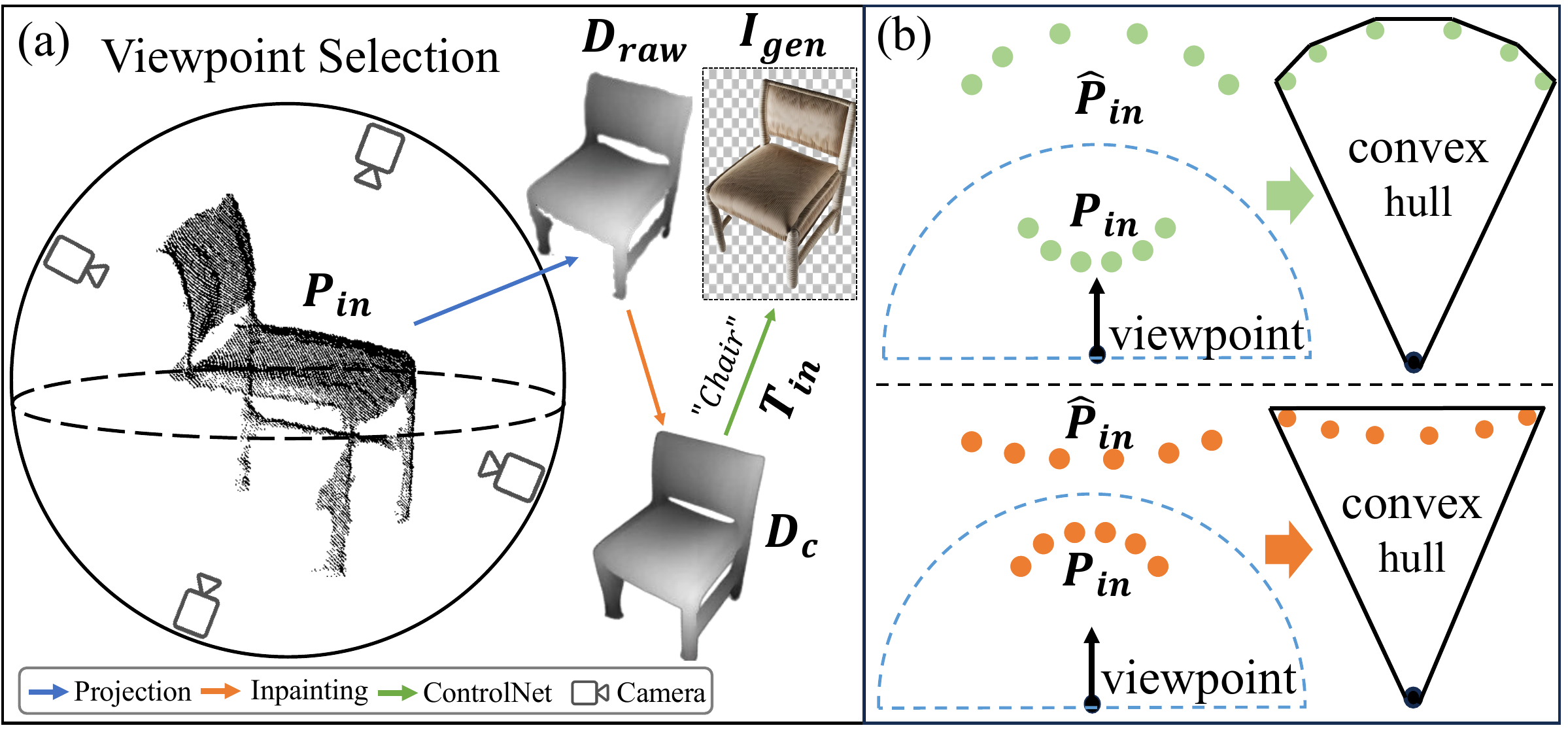}
    \caption{Illustration of Depth Prompting. (a) Overview. First, we uniformly position cameras around the partial point cloud \( P_{in} \) to select a scanning viewpoint. From this viewpoint, we project to obtain depth and the corresponding mask, and then apply mask inpainting to achieve high-quality depth. (b) Viewpoint Selection: For each viewpoint $V_i$, we perform a spherical flip on \( P_{in} \) for each camera to obtain a mirrored point cloud \( \hat{P_{in}} \), then create a convex hull around \( \hat{P_{in}} \cup V_i \), identifying the points on this hull as visible points. The camera with the greatest number of visible points is chosen as the scan viewpoint \( V_{scan} \). The top of (b) is a true viewpoint, all points lie on the convex hull. The bottom of (b) is the opposite viewpoint, only two lie on the convex hull. }
    \label{fig:depth-prompting}
\end{figure}
\label{subsec:Depth Bridging}
\subsection{Depth Prompting}
Figure~\ref{fig:depth-prompting} describes the proposed Depth Prompting module.
This module generates an RGB image from the input partial point cloud \( P_{in} \) by first projecting it to a coarse depth map $D_{raw}$ as an intermediary. Through masked inpainting of missing areas, a smooth depth map \( D_c \) is produced to enhance robustness to point cloud sparsity. Finally, \( D_c \) and the text prompt \( T_{in} \) are input into a depth-conditioning ControlNet~\cite{controlnet} to produce the corresponding RGB image.
To project a high-quality depth image from an incomplete point cloud, we propose to find the viewpoint from which the point cloud was captured. Although Huang et al.~\cite{zeroshotpointcloudcompletion} employs a distance-based method for viewpoint estimation, this approach can sometimes result in issues such as depth reversal. To address these problems, we follow the approach proposed by \cite{hiddenpointremove}, framing the viewpoint estimation as a hidden point removal task. As illustrated in Figure~\ref{fig:depth-prompting}(a), We start by evenly positioning \( M \) cameras \( V_i \) ( where \( i = 1, 2, \dots, M \) ) around the input point cloud \( P_{in} \). For each camera, as shown in Figure \ref{fig:depth-prompting}(b), we perform a spherical flip on \( P_{in} \) to obtain a mirrored point cloud \( \hat{P_{in}} \). We then create a convex hull around \( \hat{P_{in}} \cup V_i \), identifying the points on this hull as visible points. The camera with the greatest number of visible points is chosen as the scan viewpoint \( V_{scan} \). By constructing the convex hull, our approach effectively prevents depth reversal and projects $P_{in}$ to an initial depth map $D_{raw}$.

However, some partial point clouds, such as cars in the KITTI dataset, are extremely sparse, leading to sparse depth projections that hinder subsequent completion. To address this issue, we use a pre-trained 2D inpainting diffusion model~\cite{inpanting} to fill the missing holes in the sparse depth \( D_{raw} \), resulting in a complete, high-quality depth image \( D_{c} \). 
To create an inpainting mask, we first project the point cloud with a large pixel size to obtain \( M_{FULL} \). We then apply an XOR operation between \( M_{FULL} \) and the inverted depth map (\( \neg D_{raw} \)), which generates the required mask for inpainting. 
Using this mask, the inpainting model fills the missing depth regions and smooths any irregular edges, producing \( D_{c} \). Note that any inpainting model capable of filling masked areas can be applied here.
Finally, we use \( D_{c} \) as conditioning input, along with the text prompt \( T_{in} \), to generate the image \( I_{gen} \) corresponding to the partial input. This is achieved by leveraging a pre-trained depth-conditional image generation model, such as ControlNet~\cite{controlnet}.
\label{subsec:Geometric Preserving Fusion}
\subsection{Geometric Preserving Fusion}
\begin{figure*}
    \centering
    \includegraphics[width=1\linewidth]{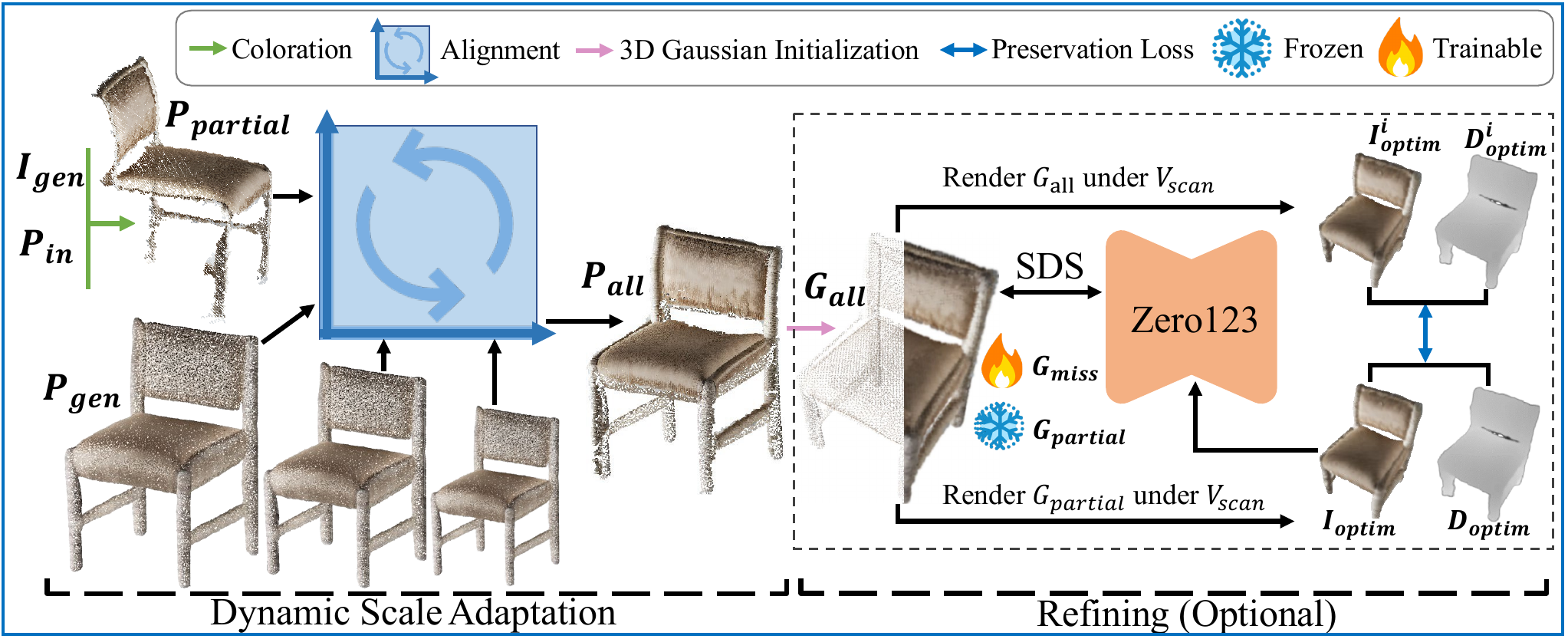}
    \caption{Illustration of Geometric Preserving Fusion. 
    In the Dynamic Scale Adaptation stage, an optimal scale factor is selected to align \( P_{partial} \) and \( P_{gen} \), producing an initial completed point cloud \( P_{all} \).
    Then, to reduce the accumulated error in the previous steps, an optional Refining operation can be performed, where \( P_{all} \) is initialized as 3D Gaussians and optimized by the SDS loss.}
    \label{fig:GPF}
\end{figure*}
In the Dynamic Scale Adaptation stage, we first colorize the input point cloud \( P_{in} \) using the generated image \( I_{gen} \), resulting in \( P_{partial} \). Then, \( P_{partial} \) and \( P_{gen} \) are aligned at dynamic scales, producing an initial, completed point cloud \( P_{all} \). 
Then, we apply an optional Refining stage. In this stage, \( P_{all} \) is initialized as 3D Gaussians \( G_{all} \), with different regions having distinct Gaussian parameter settings to preserve the original geometric details of the input point cloud while optimizing the shape of missing areas. This step helps to eliminate error accumulation and enhance overall completion quality.
\subsubsection{Dynamic Scale Adaptation}
We first use the generated image \( I_{gen} \) to obtain the 3D shape \( P_{gen} \) through the Image-to-3D generation model. Thanks to the powerful zero-shot generation performance of the pre-trained models, the generated \( I_{gen} \) and \( P_{gen} \) are highly consistent in category and shape with the input point cloud.
Next, we use \( P_{gen} \) to fill in the missing areas of the input point cloud, as shown in Figure~\ref{fig:GPF}. To improve the fusion process, we color \( P_{in} \) using the RGB information from \( I_{gen} \), creating a colored partial point cloud \( P_{partial} \). Since different parts of the object exhibit distinct colors, these colors can be regarded as semantic cues, enriching the fusion with additional contextual information for more accurate integration. Both \( P_{partial} \) and \( P_{gen} \) are then normalized to a unified scale within the range [-0.5, 0.5], reducing the search space for subsequent integration.

To eliminate the impact of both scale and pose differences, we scale \( P_{gen} \) within the range [0.8, 1.2] at intervals of 0.1, and perform ICP~\cite{icp} alignment at each scale, using the Chamfer Distance to evaluate the alignment results. We treat the color of the point cloud as semantic information, which allows us to not only supervise the alignment geometrically but also consider color information as an additional supervision signal. During the iterative registration, we calculate both the Euclidean and RGB Chamfer Distance between \( P_{partial} \) and \( P_{gen} \). The Chamfer Distance ensures accurate geometric alignment, while the RGB Chamfer Distance supervises the alignment of the semantic information, thereby improving the overall quality of the fusion. Together, they form the following objective:
% \[
% \resizebox{0.47\textwidth}{!}{ 
% $\arg \min_{P_{gen}} \left( \alpha \cdot CD_{XYZ}(P_{partial}, P_{gen}) + \beta \cdot CD_{RGB}(P_{partial}, P_{gen}) \right)$}
% \]
\resizebox{0.47\textwidth}{!}{ $\arg \min_{s \in [0.8, 1.2]} \left( \alpha \cdot CD_{XYZ}(P_{\text{partial}}, s \cdot P_{\text{gen}}) + \beta \cdot CD_{RGB}(P_{\text{partial}}, s \cdot P_{\text{gen}}) \right)$ }

% where \( \alpha \) and \( \beta \) are regularization terms, and \( s \) represents the scaling factor.
% Finally, we select the registration result that minimizes and remove points from \( P_{gen} \) that are adjacent to \( P_{partial} \) to avoid point cloud overlap, resulting in the missing portion of the point cloud \( P_{miss} \). Together, \( P_{miss} \) and \( P_{partial} \) form the preliminary complete point cloud \( P_{all} \).
where \( \alpha \) and \( \beta \) are regularization terms, and \( s \) represents the scaling factor. 
Finally, we select the registration result that minimizes the combined XYZ and RGB Chamfer distances and remove points from \( P_{\text{gen}} \) that are adjacent to \( P_{\text{partial}} \) to avoid point cloud overlap, resulting in the missing portion of the point cloud \( P_{\text{miss}} \). Together, \( P_{\text{miss}} \) and \( P_{\text{partial}} \) form the preliminary complete point cloud \( P_{\text{all}} \).
% During the registration process, \( P_{partial} \) is used as the source point cloud, which improves the registration quality but leads to a change in its position and alignment. Therefore, we need to reverse the scaling and ICP transformation matrices to restore its original pose, ultimately achieving the fused point cloud \( P_{all} \) (\( P_{all} = P_{partial} + P_{miss} \)), thus completing the preliminary point cloud completion. 
\subsubsection{Refining}
To further enhance the accuracy of point cloud completion and reduce error accumulation, we optimize the preliminarily completed point cloud, as shown in Figure~\ref{fig:GPF}. First, the point cloud is initialized as 3D Gaussians, and then distinct parameter configurations are applied to different parts of the 3D Gaussian. This approach maintains the integrity of the original part \( G_{partial} \) while optimizing the geometry of the missing part \( G_{miss} \), thereby improving the overall quality and consistency of the point cloud completion.

\noindent\textbf{Partial setup}:
For the partial point cloud \( P_{partial} \), we initialize it as a 3D Gaussians \( G_{partial} \). To preserve the original geometry, we fix parameters such as the coordinates, color, scale, and opacity, making them non-trainable. This ensures that the geometric of the partial point cloud remains unaffected during the optimization process, thereby maintaining consistency with the original input.

\noindent\textbf{Miss setup}:
For the missing point cloud \( P_{miss} \), we initialize it as a 3D Gaussians \( G_{miss} \). The scale remains fixed, as these points are uniformly sampled from the mesh surface and already have a reasonable scale. Opacity is set to 1 and remains non-trainable to ensure the stability of the Gaussian points on the surface. The color parameters are not fixed, but the learning rate is set relatively low because color carries semantic information. This allows for adjustments to the color during optimization while preserving its semantic characteristics as much as possible. The Gaussian coordinates are the main focus of the optimization, ensuring that the missing point cloud fits the shape of the partial input.

\noindent\textbf{SDS Guidance Optimization}:
Next, under the viewpoint \( V_{scan} \), we render an image \( I_{optim} \) and a depth map \( D_{optim} \) from \( G_{partial} \). We then incorporate both \( G_{miss} \) and \( G_{partial} \), and render an image \( \tilde{I}_{\mathrm{optim}}^i \) from a random viewpoint. This process is iterated multiple times, where in each iteration, we apply SDS to extract 2D priors from the pre-trained novel view synthesis diffusion model Zero123~\cite{zero123}, refining \( G_{miss} \) based on \( I_{optim} \) until satisfactory completion is achieved. The SDS Loss can be formulated as:
\[
\resizebox{0.47\textwidth}{!}{ $\nabla_{G_{all}}\mathcal{L}_{\mathrm{SDS}}=\mathbb{E}_{t,p,\epsilon}\left[(\epsilon_\phi(I_{\mathrm{optim}};t,\tilde{I}_{\mathrm{optim}}^i,\Delta p)-\epsilon)\frac{\partial I_{\mathrm{optim}}}{\partial_{G_{all}}}\right]$}
\]
where \( \epsilon_\phi(\cdot) \) is the predicted noise from the 2D diffusion prior \( \phi \), t is the time step, , $\epsilon$ is the standard noise and \( \Delta p \) represents the relative camera pose change from the scan viewpoint \( V_{scan} \), respectively. 

Additionally, to prevent other 3D Gaussians in the optimization process from affecting the geometric information of the input in the \( G_{partial} \) region, we also render images \( I_{optim}^i \) and depth maps \( D_{optim}^i \) under the viewpoint \( V_{scan} \) during the optimization iterations, and set a preservation loss \( L_{Presv} \) for the partial region:
\[
\resizebox{0.47\textwidth}{!}{ $L_{Presv} = w_1 \cdot \text{MSE}(I_{optim}, I_{optim}^i) + w_2 \cdot \text{MSE}(D_{optim}, D_{optim}^i)$}
\]
where MSE is the Mean Squared Error between the optimized and reference images \( I_{optim}^i \) and \( I_{optim} \), as well as the depth maps \( D_{optim}^i \) and \( D_{optim} \). \( w_1 \) and \( w_2 \) are weights that balance the importance of image and depth losses. By incorporating \( L_{Presv} \) and \( L_{SDS} \), our method preserves the geometry of the partial point cloud while optimizing the missing areas, reducing multi-stage error accumulation and improving the overall completion quality.

\section{Experiment}
\label{sec:experiment}

\begin{table*}[t]
    \tiny
    \renewcommand\arraystretch{1.2}
    \centering
    \caption{Quantitative results on the Redwood~\cite{redwood}~\cite{sds-complete} dataset. Ours* represents our results without Refining ({$\displaystyle \ell ^{1}$} CD and EMD $\times 10^2$ ).}
    \label{tab:redwood}
    \small
    \scalebox{0.8}{ % 缩放比例为 0.8
    \begin{tabular}{c|cccccccccc|c}
    \toprule[1pt]
    Objects & Table & Swivel-Chair & Arm-Chair & Chair & Sofa & Vase & Off-Can & Vespa & Wheelie-Bin &Tricycle & Avg$\downarrow$ \\
    
    Metrics & CD/EMD & CD/EMD & CD/EMD & CD/EMD & CD/EMD & CD/EMD & CD/EMD & CD/EMD & CD/EMD  & CD/EMD & CD/EMD  \\
    \midrule[0.3pt]
      PoinTr~\cite{Pointr}  &1.86/3.50 & 4.08/8.49 & 1.95/4.22 & 2.69/5.38 & 2.96/5.02 & 4.05/7.28 & 4.82/6.92 & 2.00/4.06 & 2.78/3.51 &1.70/3.99 & 2.89/5.24 \\
      SnowflakeNet~\cite{Snowflakenet} &3.44/6.92 & 3.40/7.58 & 2.15/4.45 & 2.35/5.28 & 2.64/5.00 & 4.63/7.69 & 4.36/6.75 & 2.07/4.42 & 3.14/5.03 &1.44/3.32 &2.96/5.64  \\
      Adapointr~\cite{AdaPoinTr}  &5.20/6.44 & 5.09/8.03 & 3.67/4.53 & 4.40/5.96 & 3.59/5.18 & 6.23/7.56 & 6.04/7.69 & 3.21/4.65 & 4.13/7.63 &2.90/4.25 &4.45/6.19 \\
      SDS-Complete~\cite{sds-complete}  &1.67/2.92 & 2.24/3.09 & 2.18/3.16 & 2.62/3.61 & 2.95/4.56 & 3.26/5.89 & 4.03/4.36 & 3.46/5.94 & 2.69/3.21 &2.11/3.87 &2.72/4.06 \\
      \midrule[0.3pt]
      Ours* &1.41/2.24 & 1.69/2.37 & 1.38/1.76 & 1.47/2.48 & 1.61/2.93 & 3.15/5.24 & 3.04/4.62 & 1.59/2.83 & 2.65/3.64 &1.79/3.52 &1.98/3.16  \\
      Ours &\textbf{1.28/2.07} & \textbf{1.43/2.29} & \textbf{1.16/1.68} & \textbf{1.36/2.20} & \textbf{1.58/2.78} & \textbf{2.86/4.85} & \textbf{2.72/4.36} & \textbf{1.36/2.47} & \textbf{2.31}/\textbf{3.17} &\textbf{1.38}/\textbf{2.97} & \textbf{1.74}/\textbf{2.88} \\
      \bottomrule[1pt]
    \end{tabular}
    }
\end{table*}

\begin{table}[t]
        \renewcommand\arraystretch{1.2}
        \centering
        \caption{Quantitative results on ScanNet ($\ell ^1$ CD and EMD $\times 10^2$).}
        \label{tab:scannet}
        \small
        \begin{tabular}{c|c c| cc} 
        \toprule[1pt]
        Methods& SnowflakeNet & AdaPoinTr & Ours \\
        Metrics & CD/EMD & CD/EMD & CD/EMD \\
         \midrule[0.3pt]
         Table& 1.80/4.78 & 2.34/6.12 & \textbf{1.67}/\textbf{3.86}\\
         Chair& 1.68/3.76 & 2.07/\textbf{3.09} & \textbf{1.57}/3.24\\
         Avg & 1.74/4.27 & 2.21/2.38 & \textbf{1.62}/\textbf{3.55}\\
        \bottomrule[1pt]
        \end{tabular}
\end{table}

\begin{table}[t]
        \renewcommand\arraystretch{1.2}
        \centering
        \caption{Performance of ablation variant C (w/o depth inpainting) on different datasets. Variant C performs relatively well on Redwood's dense point clouds but shows significant performance drops with the sparse point clouds in ScanNet ({$\displaystyle \ell ^{1}$} CD and EMD $\times 10^2$).}
        \label{tab:ablationDepthInpainting}
        \small
        \begin{tabular}{c|c c }
        \toprule[1pt]
        Methods     & variant C & Ours\\
        Metrics     & CD/EMD & CD/EMD\\
        \midrule[0.3pt]
        Redwood     &2.23/3.60 & \textbf{1.74}/\textbf{2.88} \\
        ScanNet     &3.57/6.10 & \textbf{1.62}/\textbf{3.55}  \\
        \bottomrule[1pt]
        \end{tabular}
\end{table}

\begin{table}[t]
    \renewcommand\arraystretch{1.2}
    \centering
    \caption{Performance of ablation variants on the Redwood dataset ($\ell ^1$ CD and EMD $\times 10^2$).}
    \label{tab:ablationVariants}
    \small
    \begin{tabular}{c|c c }
    \toprule[1pt]
     Methods  & CD$\downarrow$ & EMD$\downarrow$   \\
    \midrule[0.3pt]
    A : w/o Viewpoint Selection  & 2.44 & 3.79 \\
    B : w/o ControlNet & 4.31 & 6.80 \\
    C : w/o Depth Inpainting &2.23 & 3.60 \\
    D : w/o 3D Generative Model & 4.65 & 6.13 \\
    E : w/o Dynamic Scale Adaptation & 4.38 & 4.52 \\
    F : w/o SDS Optimization & 1.98 & 3.16 \\
    Ours  & \textbf{1.74} & \textbf{2.88} \\
    \bottomrule[1pt]
    \end{tabular}
\end{table}

% \begin{table}[t]
%         \renewcommand\arraystretch{1.2}
%         \centering
%         \caption{Comparison of Completion Time on the Redwood Dataset. Ours* represent without using SDS optimization.}
%         \label{tab:ablationtime}
%         \normalsize
%         \begin{tabular}{c|c c c }
%         \toprule[1pt]
%         Methods  & CD$\downarrow$ & EMD$\downarrow$  & Time$\downarrow$ \\
%         \midrule[0.3pt]
%         SDS-Complete &2.72 & 4.06 & 40Hours \\
%         Ours* & 1.98 &3.16 & \textbf{1 Minute}\\
%         Ours & \textbf{1.74} &\textbf{2.88} & 2 Minutes\\
%         \bottomrule[1pt]
%         \end{tabular}
% \end{table}

\begin{figure*}
    \centering
    \includegraphics[width=1\textwidth]{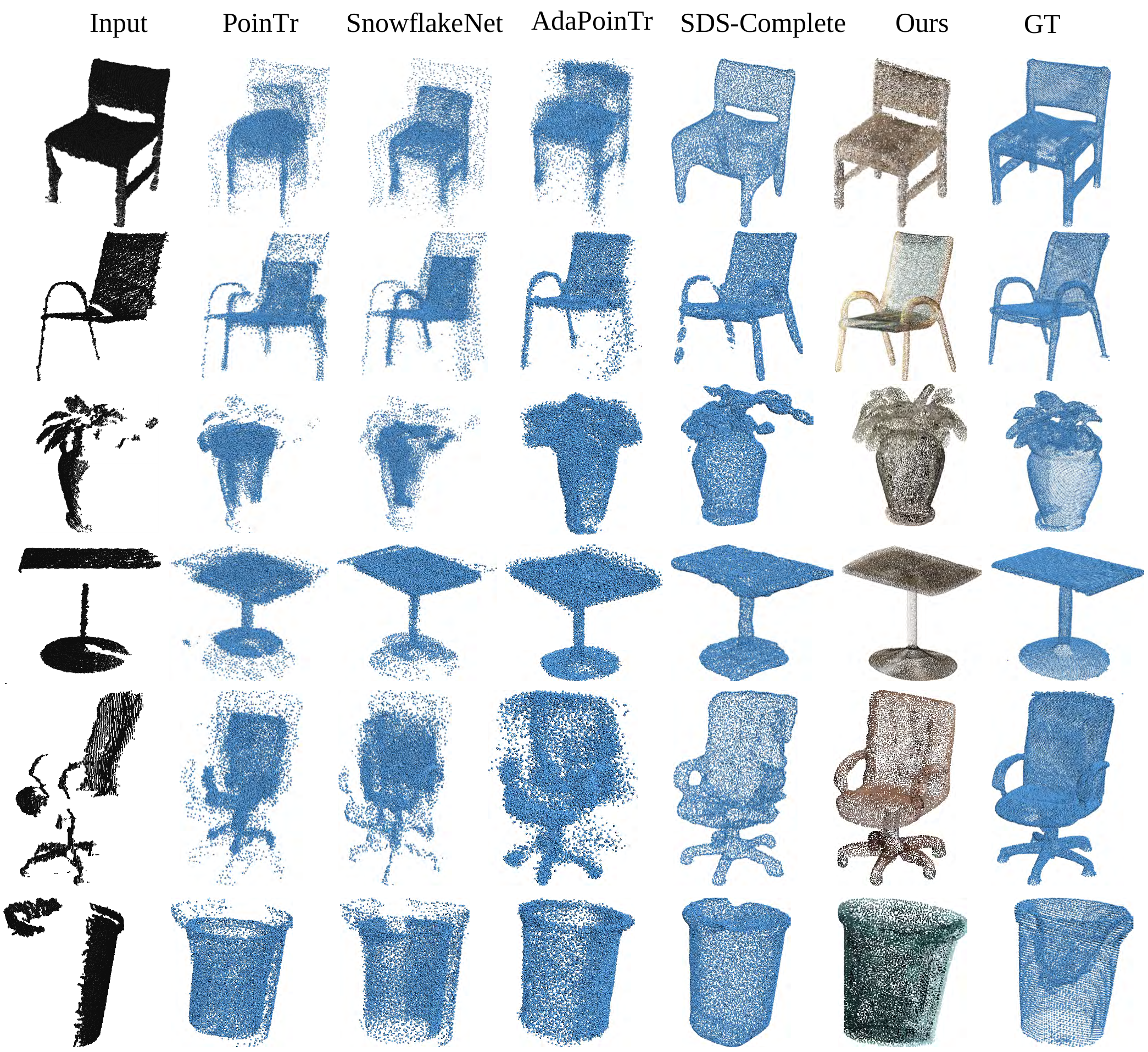}
    \caption{Visual comparisons with recent methods~\cite{Pointr,Snowflakenet,AdaPoinTr} on the Redwood dataset.}
    \label{fig:exp_redwood}
\end{figure*}

\begin{figure}
    \centering
    \includegraphics[width=1\linewidth]{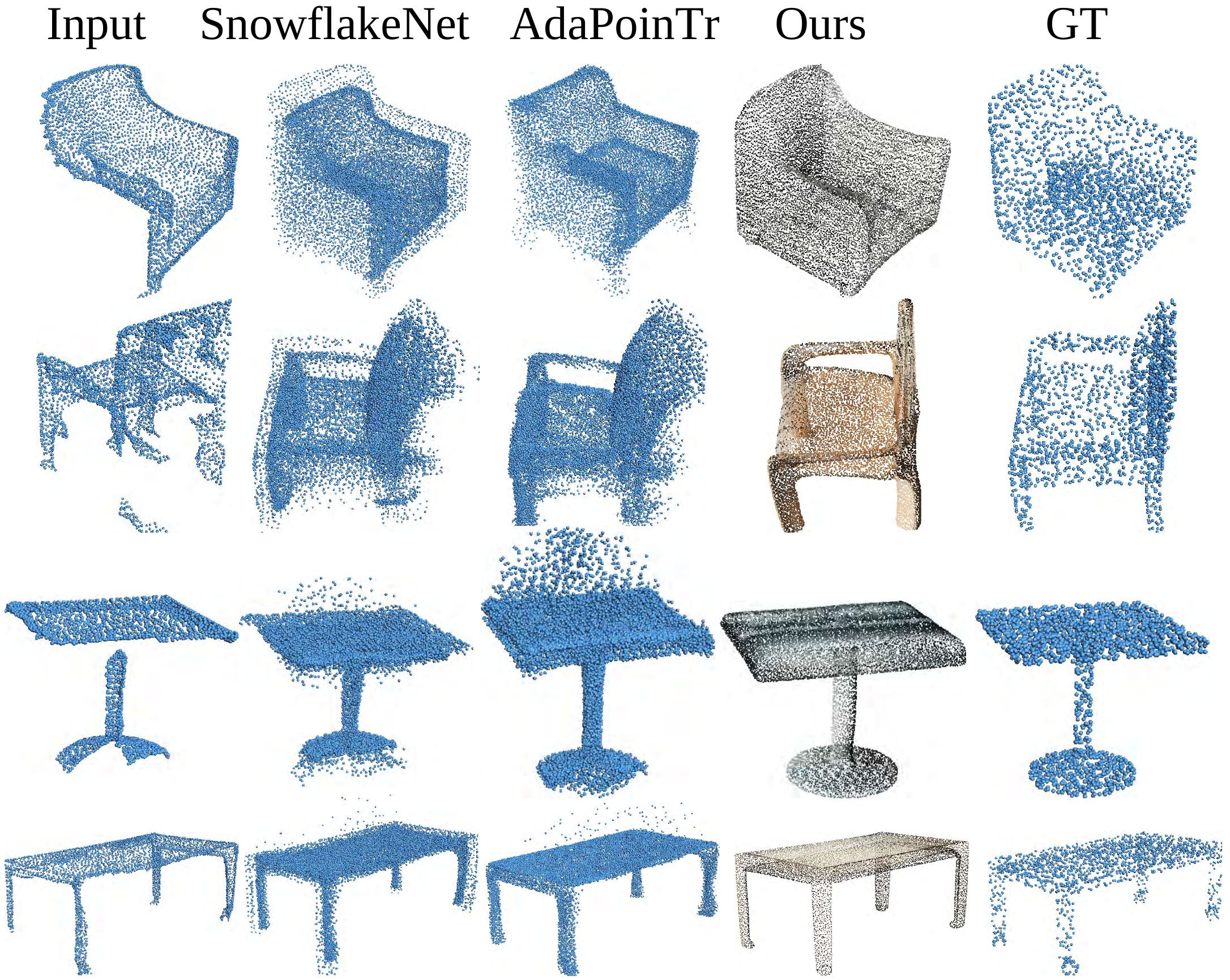}
    \caption{Visual comparisons with recent methods ~\cite{Snowflakenet,AdaPoinTr} on the ScanNet dataset.}
    \label{fig:exp_scannet}
\end{figure}

\begin{figure}
    \centering
    \includegraphics[width=\linewidth]{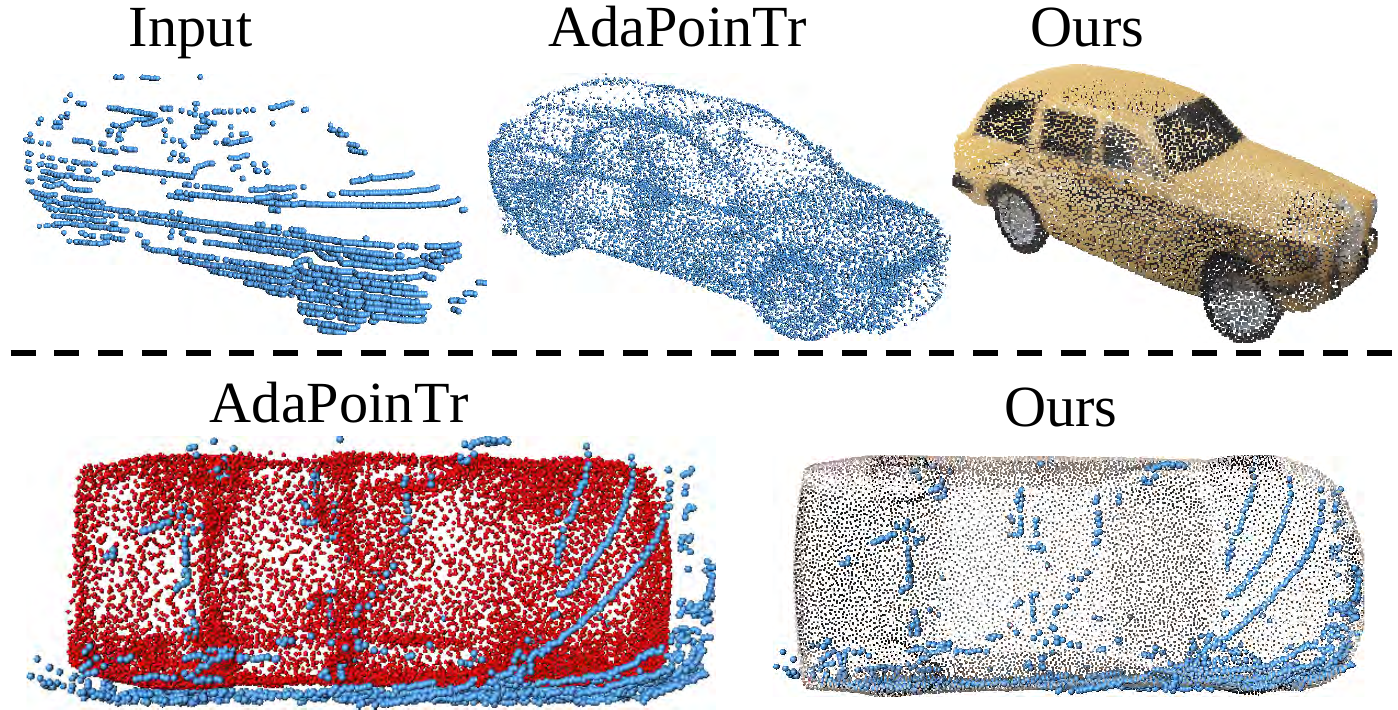}
    \caption{(Top) Visual comparisons with AdaPoinTr~\cite{AdaPoinTr} on KITTI~\cite{kitti}. (Bottom) We display the point clouds in different colors: blue for the Partial Input, red for AdaPoinTr, and gray for Ours. Our result maintains a consistent scale with the input.}
    \label{fig:exp_kitti}
\end{figure}

\begin{figure}
    \centering
    \includegraphics[width=1\linewidth]{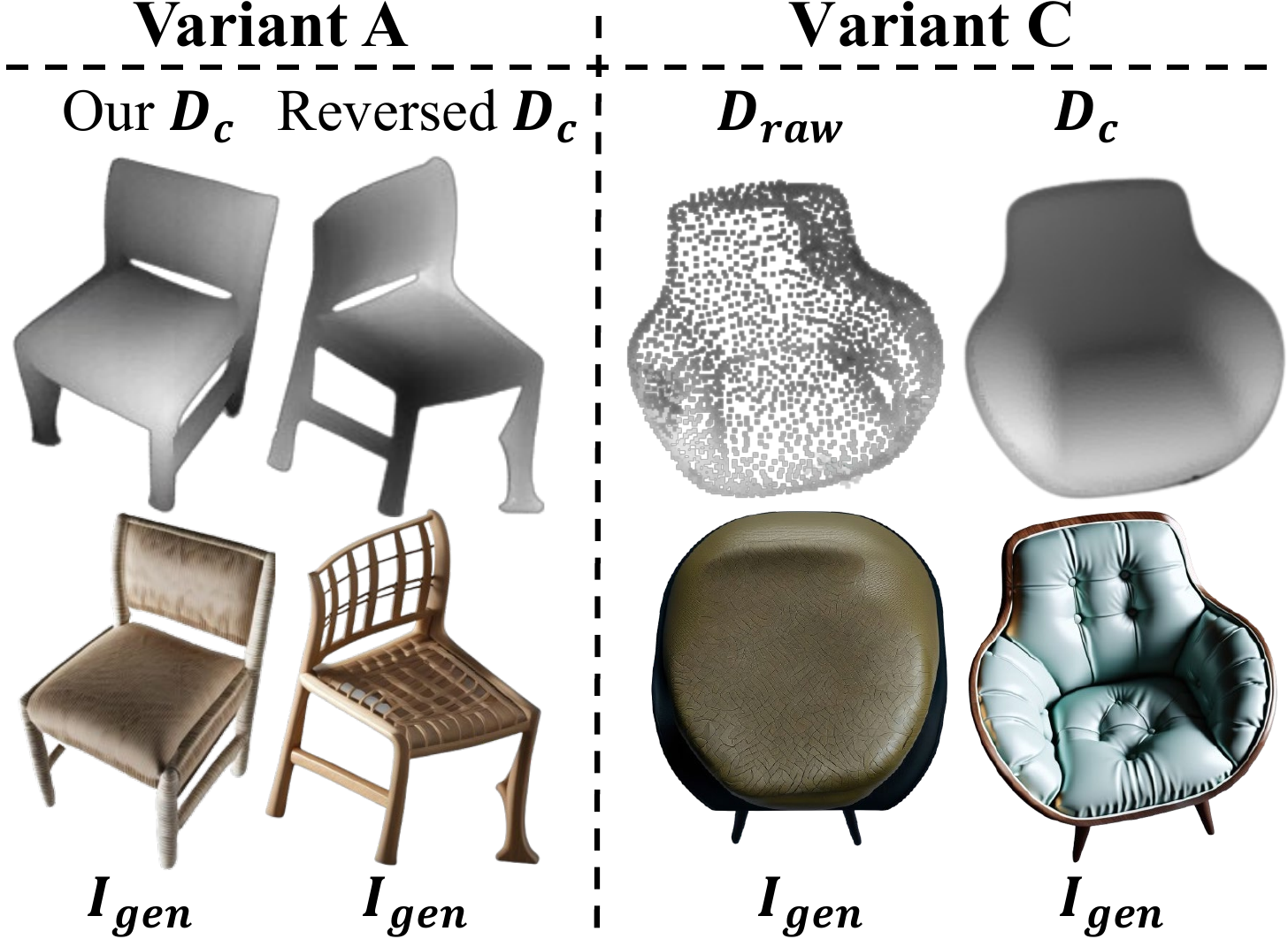}
    \caption{Depth and RGB images produced by variant A (w/o Viewpoint Selection) and Variant C (w/o Depth Inpainting). (Left) The above is the complete depth \( D_c \) obtained from the scanning viewpoint and the opposite viewpoint, and below are the corresponding generated images $I_{gen}$. (Right) The above are the sparse depth \( D_{\text{raw}} \) and the complete depth \( D_{\text{c}} \), and below are the generated corresponding images \( I_{\text{gen}} \).}
    \label{fig:exp_abs}
\end{figure}

% \subsection{Implementation Details}
% % 这里写两个2D diffusion的参数，比如推理步数、guidance scale等
% % 师兄， inpainting 用的是Diffusion Models Beat GANs on Image Synthesis这个论文。
% For depth inpainting, we employ a pre-trained diffusion model~\cite{inpanting} with a resolution of 256×256. For the depth-conditioned generative model, we utilize ControlNet~\cite{controlnet} with inference steps set to 30 and a conditioning scale of 0.99, generating outputs at a resolution of 1024×1024. For the Image-to-3D generative model, InstantMesh~\cite{instantmesh}, we use a base-scale configuration with inference steps set to 75, scale set to 1, distance parameter at 4.5, and six views for rendering.
% During SDS optimization, the output resolution is set to 256×256 with a batch size of 12.

\subsection{Dataset and Evaluation Metric}
We validate our method on three real-world datasets Redwood~\cite{redwood}, ScanNet~\cite{dai2017scannet}, and KITTI~\cite{kitti}.
For the Redwood~\cite{redwood} dataset, we follow prior approaches~\cite{sds-complete}, using single-view scans as partial inputs and multi-frame aggregations as ground truth. Since previous deep learning-based methods were trained on standardized synthetic datasets, we normalize the Redwood dataset point clouds to the range [-0.5, 0.5] and set the elevation angle to 0° to ensure a fair comparison with their input requirements.
For the ScanNet dataset, which contains partial point clouds extracted from RGB-D scans, we focused on tables and chairs due to their complex structures and additional supports that introduce challenging self-occlusion cases for our method. For each category, we select 16 objects for validation. In addition, we used the ground truth provided by~\cite{wu2023scoda}, consisting of 2048 points for quantitative evaluation.

For quantitative evaluation, we followed prior methods by sampling 16,384 points from the Redwood dataset and 2,048 points from the ScanNet dataset using Farthest Point Sampling (FPS) to enable direct comparison with the ground truth. To assess the quality of point cloud completion, we used the widely adopted Chamfer Distance (CD) and Earth Mover’s Distance (EMD) metrics, scaling the loss values by a factor of 100 for clearer interpretation.
We also conduct a qualitative evaluation on KITTI~\cite{kitti} to assess the performance on sparse LiDAR scans.

\subsection{Results on the Redwood dataset}
The quantitative and qualitative results are presented in Table~\ref{tab:redwood} and Figure~\ref{fig:exp_redwood}. With or without the SDS Refining step, GenPC consistently achieves state-of-the-art performance across the entire dataset. These results indicate that existing learning-based methods~\cite{Pointr, Snowflakenet, AdaPoinTr} struggle to complete out-of-distribution data, even when these data belong to categories seen during training (e.g., chairs and couches). Additionally, these methods are sensitive to scale variations, leading to inconsistent outputs when the input scale changes. Compared with the only zero-shot method, SDS-Complete~\cite{sds-complete}, GenPC achieves an average reduction in CD by $36\%$ and EMD by $29\%$. Furthermore, Figure~\ref{fig:exp_redwood} clearly illustrates that GenPC outputs finer structure details than SDS-Complete, attributed to the rich geometric priors provided by the pre-trained 3D generative model.

\subsection{Results on the ScanNet dataset}
Comparison with two cutting-edge learning-based methods~\cite{Snowflakenet,AdaPoinTr} are presented in Table~\ref{tab:scannet} and Figure~\ref{fig:exp_scannet}. 
Our method demonstrates advanced performance in completion quality, maintaining reliable results even when dealing with sparse and noisy point clouds. As shown in Figure~\ref{fig:exp_scannet}, our method generates completion outputs with high fidelity to the input point cloud and rich geometric details, while learning-based methods are affected by domain gaps, leading to numerous noisy points in their results. 
% Considering that SDS-Complete~\cite{sds-complete} requires up to 40 hours to complete a single object, we did not include it in our evaluations.

\subsection{Results on KITTI}

A qualitative comparison on the KITTI dataset is presented in Figure~\ref{fig:exp_kitti}, which shows that GenPC produces results with a complete and realistic shape without any extraneous noise. In contrast, previous methods trained on ShapeNet produce completed point clouds that are smaller in scale than the original, as shown in the bottom of \ref{fig:exp_kitti}. The proposed Dynamic Scale Adaptation allows the completed results to maintain scale consistency with the original point cloud.
 
% \subsection{Results on the synthetic Dataset}
% To validate the effectiveness of our method on synthetic datasets, we conducted tests on the chair category within the sparse point cloud PCN test set. Since chairs typically have multiple supporting structures (such as legs), they exhibit more complex self-occlusion compared to other categories. As a result, both the depth bridging and fusion module in our method face greater challenges.

\subsection{Ablation Study}
% We remove and modify the main components to ablate GenPC. All ablation variants are tested on the real-world datasets. The ablation variants can be categorized as ablations on Depth Prompting module, image-to-3D generative model and Geometric Preserving Fusion module. 
% Since our approach is multi-stage, later inputs heavily depend on earlier outputs, making certain ablation experiments infeasible. Therefore, in some cases, we only perform ablation at a specific stage.
\subsubsection{Ablation on Depth Prompting Module}
To investigate the impact of the depth extraction method, we compare three variants of Depth Prompting. 
In variant A, we replace our viewpoint selection with a distance-based method similar to \cite{zeroshotpointcloudcompletion}, leading to significantly increased CD and EMD values. 
Meanwhile, as shown in Figure~\ref{fig:exp_abs}, although this method correctly identifies the viewpoint in some cases, it may select the reverse viewpoint, causing depth flipping. This flipped depth map disrupts accurate image generation and severely impacts completion quality.
In variant B, ControlNet is removed, and the inpainted depth $D_c$ is used as input to the image-to-3D generative model to examine the effects of color information on subsequent processes. In some cases, experimental observations show that, even with high-quality depth, the generated 3D shapes are reasonable but lack color, rendering them unsuitable for SDS optimization in the second stage. 
In variant C, we skip the depth inpainting step to evaluate the effect of low-quality depth on downstream processes. As shown in Figure~\ref{fig:exp_abs}, depth maps projected from sparse point clouds fail to generate accurate images, resulting in a significant drop in performance. Therefore, while this variant performs well on dense point cloud datasets like Redwood, it struggles on sparse point cloud datasets like ScanNet, as shown in Table~\ref{tab:ablationDepthInpainting}. 

\subsubsection{Ablation on 3D Generative Model}
To examine the effect of the image-to-3D generative model in our pipeline, we form variant D by replacing the generated 3D shape with a set of Gaussian noise point clouds. 
% These noise point clouds are initialized as 3D Gaussians alongside the partial input, maintaining the same 3D Gaussian parameter settings as in Geometric Preserving Fusion, except for the learning rate of the color parameters. 
The Refine step is then applied, optimizing over 5000 iterations in an attempt to complete the missing regions. 
The results in Table~\ref{tab:ablationVariants}, the absence of explicit geometric priors significantly impacts the completion performance. 

\subsubsection{Ablation on Geometric Preserving Fusion Module}
In variant E, we directly align the generated 3D shape \( P_{\text{gen}} \) with \( P_{\text{partial}} \) without using Dynamic Scale Adaptation to validate the effectiveness of this process. Due to scale inconsistency, the direct alignment fails to properly match the two point clouds, thereby wasting the rich geometric priors provided by the 3D shape.
In variant F, we omit the Refining process and use the merged point cloud \( P_{\text{all}} \) directly as the completion result. While quantitative metrics show that the Refining process can further enhance the overall completion quality, our experiments reveal that the merged point cloud \( P_{\text{all}} \) often performs competitively in both visualization and quantitative metrics. Therefore, we make the Refining process optional to improve completion speed.

\section{Conclusion}
\label{sec:conclusion}
In this study, we make the first attempt to leverage a pre-trained 3D generative model for zero-shot point cloud completion and introduce GenPC.
To capitalize on the generative model’s inherent generalization ability, our framework consists of two key components: Depth Prompting and Geometric-Preserving Fusion.
The Depth Prompting module prompts an image-to-3D generative model with the partial point cloud. Then, the Geometric Preserving Fusion module aligns the partial input with the generated 3D shape by dynamically adjusting its pose and scale. 
Experiments on widely used datasets demonstrate that GenPC achieves state-of-the-art performance. With the explicit geometric prior from the 3D generative model, GenPC takes a step closer towards robust real-world scan completion.

{\small
\bibliographystyle{ieeenat_fullname}
\bibliography{11_references}
}

\ifarxiv \clearpage \appendix % \section{Appendix Section}
% \label{sec:appendix_section}
% Supplementary material goes here.
 \fi

\end{document}